\pdfoutput=1
\documentclass[11pt]{article}

\usepackage[final]{acl}

\usepackage{times}
\usepackage{latexsym}
\usepackage{tcolorbox}
\usepackage{multicol} % for multicolumns in tables

\usepackage[T1]{fontenc}
\usepackage[utf8]{inputenc}

\usepackage{microtype}

\usepackage{inconsolata}

\usepackage{blindtext}
\usepackage{hyperref}
\usepackage{graphicx}
\usepackage[textsize=tiny]{todonotes}
\usepackage{booktabs}
\usepackage{makecell}
\usepackage{amsmath}
\usepackage{footnote}

\title{DENIAHL: In-Context Features \\ Influence LLM Needle-In-A-Haystack Abilities}

\author{Hui Dai\textsuperscript{*,1}, 
Dan Pechi\textsuperscript{*,1,2},
Xinyi Yang\textsuperscript{1}, 
Garvit Banga\textsuperscript{1}, Raghav Mantri\textsuperscript{1} \\
  \textsuperscript{1}New York University \\
  \textsuperscript{2}Databricks \\
  \textsuperscript{*} Equal Contribution \\
  \texttt{\{hd2584, danpechi,	xy2587, gb2762, rm6418\}@nyu.edu}
}

\begin{document}
\maketitle
\begin{abstract}

The Needle-in-a-haystack (NIAH) test is a general task used to assess language models' (LMs') abilities to recall particular information from long input context. This framework however does not provide a means of analyzing what factors, beyond context length, contribute to LMs' abilities or inabilities to separate and recall needles from their haystacks. To provide a systematic means of assessing what features contribute to LMs' NIAH capabilities, we developed a synthetic benchmark called DENIAHL (Data-oriented Evaluation of NIAH for LLM's). Our work expands on previous NIAH studies by ablating NIAH features beyond typical context length including data type, size, and patterns. We find stark differences between GPT-3.5 and LLaMA 2-7B's performance on DENIAHL, and drops in recall performance when features like item size are increased, and to some degree when data type is changed from numbers to letters. This has implications for increasingly large context models, demonstrating factors beyond item-number impact NIAH capabilities. \\
\href{https://github.com/ameliadai/DENIAHL}{https://github.com/ameliadai/DENIAHL}

\end{abstract}

\begin{figure}[ht!]
\centering
\includegraphics[width=75mm]{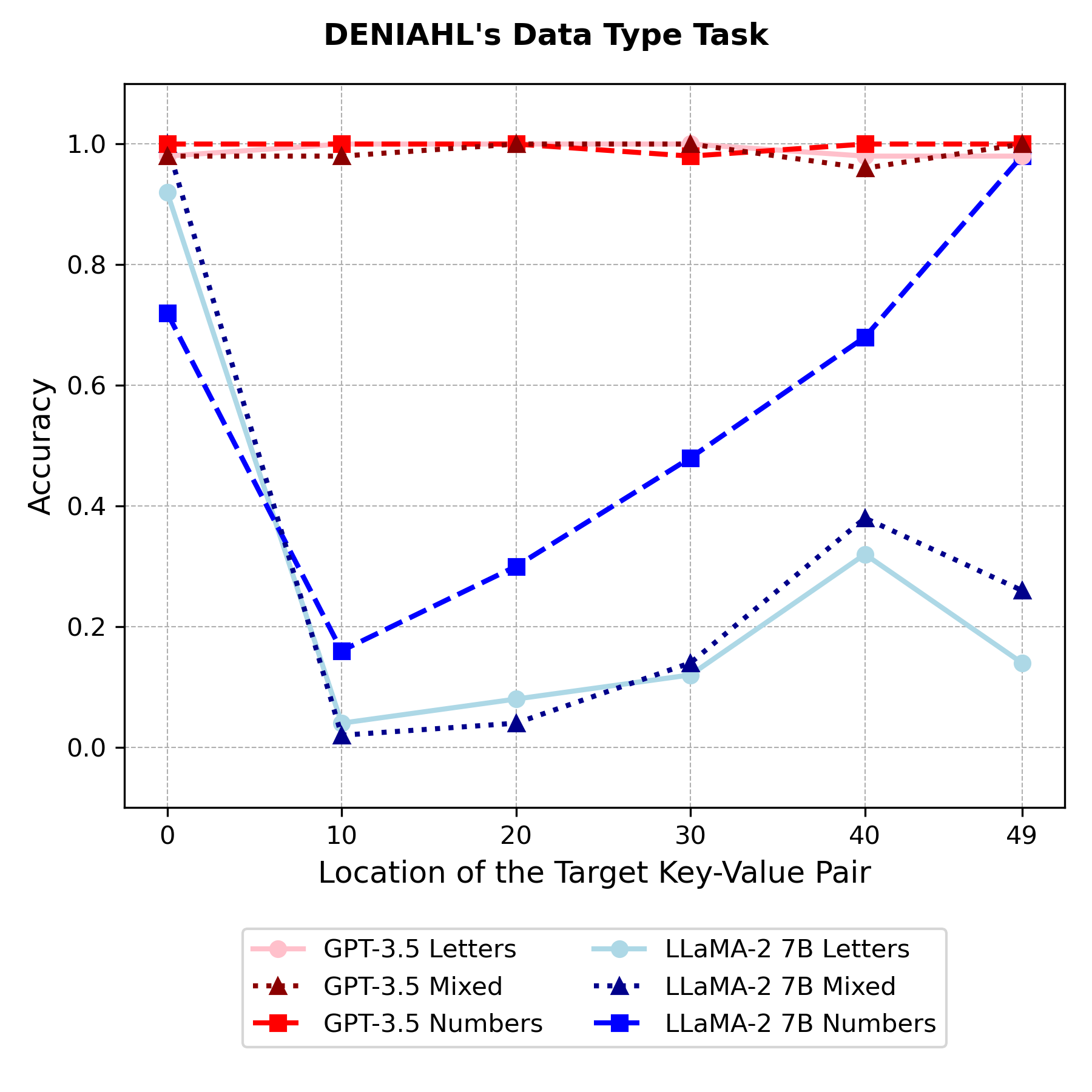}
\caption{In \textbf{Data-oriented Evaluation of Needle in A Haystack for LLM's (DENIAHL)}, we demonstrate properties of LLMs' context data strongly influence performance on Needle-in-a-haystack tasks. In addition to showing more robust performance in GPT-3.5 vs LLaMA-2 7B, we find that context containing numbers present typical "lost-in-the-middle" phenomena, whereas letter data is also poorly recalled at the end of the LLM's input context.}
% \caption {Data type influences "lost-in-the-middle" versus "lost-in-the-end" phenomena}
\label{fig: item_type}
% \label{figure: main-fig-num-let}
\end{figure}

\section{Introduction}

In the past few years, large language models (LLMs) have developed increasingly large context sizes \cite{bulatov2024scaling, ding2024longrope, xiong2023effective}. More recent models even report having infinite context size \cite{munkhdalai2024leave}. These models are often presented as alternatives to compound, retrieval-augmented generation systems that can retrieve information from multi-page documents using similarity search \cite{xu2024retrieval}.

\begin{table*}[ht]
\centering
\resizebox{\linewidth}{!}{
\begin{tabular}{@{}lllll@{}}
\toprule
\textbf{Task} & \textbf{Configuration} & \textbf{Example} & \textbf{Key} & \textbf{Gold Label} \\
\midrule
\begin{minipage}[t]{0.12\linewidth}
Data Size
\end{minipage} &
\begin{minipage}[t]{0.25\linewidth}
Number of items: varying the number of key-value pairs
\end{minipage} &
\begin{minipage}[t]{0.5\linewidth}
\textit{number of items = 3} \\
\{ "e3c90162-c33e-46": "113007f3-4421-42", \\
 \textbf{"05b17449-c3d6-4b": "76231614-9b70-4f"}, \\
 "bb7a3f7e-55a7-45": "22ef5b80-1656-40"\}
\end{minipage} &
\begin{minipage}[t]{0.1\linewidth}
05b17449-c3d6-4b
\end{minipage} &
\begin{minipage}[t]{0.1\linewidth}
76231614-9b70-4f
\end{minipage}
\\
\begin{minipage}[t]{0.1\linewidth}
\end{minipage} &

\begin{minipage}[t]{0.25\linewidth}
Item length: varying the length of the keys/values
\end{minipage} &
\begin{minipage}[t]{0.5\linewidth}
\textit{item length = 8} \\
\{ \textbf{"66b1b1c9": "d53e0cf5"},\\ 
            ... \\
            "6669d056": "cae5c7ee" \}
\end{minipage} &
\begin{minipage}[t]{0.1\linewidth}
66b1b1c9
\end{minipage} &
\begin{minipage}[t]{0.1\linewidth}
d53e0cf5
\end{minipage}
\\
\midrule
\begin{minipage}[t]{0.12\linewidth}
Pattern
\end{minipage} &
\begin{minipage}[t]{0.25\linewidth}
Numerical pattern \\
Letter pattern 
\end{minipage} &
\begin{minipage}[t]{0.5\linewidth}
\{ 1: 3, 2: 5, \textbf{3: 6}, 4: 9, 5: 11\}\\
\{ "xf": "XF",
 "gf": "GF",
\textbf{ "pr": "AB"},
 "pl": "PL"\} \\
*Note that we break the pattern of the value for the corresponding key we want to retrieve
\end{minipage} &
\begin{minipage}[t]{0.1\linewidth}
3 \\
pr\\
\end{minipage} &
\begin{minipage}[t]{0.1\linewidth}
6 \\
AB\\
\end{minipage}
\\
\midrule
\begin{minipage}[t]{0.12\linewidth}
Data Type
\end{minipage} &
\begin{minipage}[t]{0.25\linewidth}
All letters \\
All numbers \\
Mixed 
\end{minipage} &
\begin{minipage}[t]{0.5\linewidth}
\{ "aBc": "xYz", "Aaa": "sdD", \textbf{"xcL": "AQw"} \}\\
\{ 079: 234, \textbf{194: 566}, 243: 746  \}\\
\{ \textbf{"a03xcl": "fef4t5"}, "Qs235r": "2ref63"\}
\end{minipage} &
\begin{minipage}[t]{0.1\linewidth}
xcL\\
194\\
a03xcl
\end{minipage} &
\begin{minipage}[t]{0.1\linewidth}
AQw\\
566\\
fef4t5
\end{minipage}
\\
\bottomrule
\end{tabular}}
\caption{DENIAHL task categories with examples. In addition to the 3 categories we propose, we also vary the number of items as specified in row 2, similar to other benchmarks.}
\label{tab:example-task}
\end{table*}

While convenient to use a single model, even SoTA LLM's with billions of parameters demonstrate an inability to recall information from arbitrary positions in that context \cite{anthropic2024claude}. \citet{liu2023lost} formalized an evaluation for this context recall task, referred to as the Needle in a Haystack (NIAH) test. A NIAH test typically involves passing long context (the haystack), along with specific information in that context (the needle) that the LLM is expected to recall. One may observe poor NIAH performance when passing the longest possible input to a particular LLM, effectively increasing the amount of hay in the haystack. NIAH tests can involve either linguistic data as in \citet{niah}, or Key-Value pairs \cite{niah}. However, different features of that input presumably influence performance. 

While several studies have looked at factors like the number of tokens in input \cite{levy2024task}, we hypothesize that beyond input size, other factors like data type, item length, and patterns can influence LLMs' recall ability, just as they influence humans' recall. For example, the sequence "123456789" is far easier to recall than the same numbers ordered as "938412675". We would suspect for NIAH, the first sequence would be more easily recalled as the sequence is higher probability, similar to humans. 

In this work, we propose Data-centric Evaluation of NIAH for LLM's (DENIAHL), a new benchmark to evaluate data features that influence long-context modeling capabilities in LMs. In DENIAHL, we assess three task categories to test how data features influence recall: data size, patterns, and type. We assess LLaMA-2 7B and GPT-3.5 on DENIAHL, and report these models' performance on other common, context-truncated NIAH benchmarks for comparison.

In DENIAHL, we find LLaMA-2 7B struggles with most NIAH tests compared to GPT-3.5, with GPT-3.5 performance only degrading when presented with long, mixed-type data. We also observe a mix of lost-in-the-middle and lost-in-the-end phenomena, number-type data presenting the former, and thus generally higher recall than letter-type data. Lastly, we demonstrate that LLaMA-2 7B actually performs better with broken patterns than consistent patterns in some tasks. 

These findings underscore that NIAH performance is data-dependent. Models exhibit strong recency biases independent of data, and may not leverage global attention patterns when recalling information from context. Thus, it is important to assess these data features even when LM's have sufficient context size.

\section{Related Work}
%TODO: add more here - can steal from RULER
\subsection{Long-context Language Models}
Models have grown increasingly large \cite{hoffmann2022training}. However, as of late, more and more models are also being developed with larger, and even infinite context token sizes \cite{bulatov2024scaling, munkhdalai2024leave, mohtashami2023landmark}. ChatGPT, initially developed with a input size of 4096 tokens, now has variants of 32k token input size, demonstrating the demand for these large context models. 

\subsection{Long-context Benchmarks and Features}
As these larger context size models gain traction, it has becoming increasingly important to evaluate these models. Many of these long-context models have already been exposed for lacking the ability to use all their context, including for in-context learning \cite{li2024longcontext}. \citet{pang2022quality} introduced some of the earliest benchmarks for long-context models with a question-answering dataset with an average length of 5000 tokens. The concept of Needle-in-a-haystack and the lost-in-the-middle phenomenon added to the notion of data structure influencing recall performance \cite{liu2023lost}. Other popular NIAH and long-context benchmarks include LongBench \cite{bai2023longbench}, which includes multilingual QA benchmarks. 

Ruler \citep{hsieh2024ruler} is most similar to our work, focusing on synthetic key-value data benchmarks for NIAH. However, the authors focus on multiple needles, and other patterns. Others have also examined attention heads as a means of mechanistic interpretability for long-context, finding different heads relevant for certain components of recall \cite{wu2024retrieval}. \citet{machlab2024llm} additionally finds NIAH performance influenced by biases in models' training data.

\section{Data-centric Evaluation of Needle-in-a-haystack for LLM's}
DENIAHL consists of tasks which vary 3 categories of data properties, namely the data's \textit{size, patterns} and  \textit{type} (ie numbers vs letters) within the model's input. Contrary to current benchmarks which mostly assess data size as measured by the length of the context input, through various key-value retrieval tasks, DENIAHL tests the models' recall capabilities when data items vary according to a broader range of relevant data properties.

\subsection{Problem Definition}
In DENIAHL, we manipulate the length of input context data, varying both the length by changing the number of key-value pairs, as well as the length of the keys and values. Additionally, we assess NIAH performance when the data follows patterns, and when that pattern is broken, giving a sense of whether the model is performing fine-grained recall or merely attending to a global pattern. Lastly, we assess data type by comparing recall abilities for key-values consisting of random characters, numbers, and a mix of both. We test both LLaMA-2 7B and GPT-3.5 on DENIAHL, varying pattern, size, and type features of input data to assess how these features impact NIAH performance, and whether previously-reported lost-in-the-middle effects persist.
\subsection{Datasets}
\subsubsection{Benchmarks}
To evaluate the ability of different language models to recall input information against existing benchmarks, we evaluate LLaMA-2 7B and GPT-3.5 models on Paul Graham's essays, as used in the Needle-in-a-haystack tests \cite{niah}. This dataset consists of text data where needles are sentences that include the answer for the query. We also benchmark against the key-value datasets introduced by \citet{liu2023lost}, where the model retrieves specific values from different indices in a JSON-like input.

\subsubsection{DENIAHL datasets}
We propose a new benchmark which consists of custom datasets with varying properties in key-value formats. 
Examples and descriptions of our custom dataset's three task categories are provided in Table \ref{tab:example-task}. The task categories are as follows:
\\[0.1cm]
\textbf{Data Size}: \\
$\bullet$ With a fixed length of each key-value item, we vary context sizes based on the \textbf{number of key-value pairs} in the input prompt. \\
$\bullet$ With a fixed number of key-value pairs in each dataset, we vary context sizes based on the \textbf{length of key-values}. \\[0.1cm]
\textbf{Pattern}: \\
The key-value pairs in the pattern category follow certain \textbf{numerical} or \textbf{alphabetical} patterns. To evaluate the model's ability for accurate retrieval versus mere pattern recognition, we break the pattern at a certain position by altering the value, and observe the model's output when it is asked to retrieve that value. For example, consider a dataset having an arithmetic sequence pattern with keys (1, 2, 3, 4, 5) and corresponding values (3, 5, 7, 9, 11). If we alter the third value from 7 to 6, we can assess whether the model accurately retrieves the altered value (6) or erroneously predicts the original sequence value (7), thus revealing the model's reliance on pattern inference. \\[0.1cm]
\textbf{Data Type}: \\
We vary keys and values as random \textbf{numbers}, \textbf{letters}, or a \textbf{mixture} of both. \\
Varying these data properties of key-value pairs, we test how model performance changes with respect to each variable.

\section{Experimental Setup}
The models we test are LLaMA-2 7B and GPT-3.5. Following the framework established by \citet{liu2023lost}, we define the key-value retrieval task as follows: Given a string-serialized JSON object with $k$ key-value pairs and a query requesting the retrieval of $\text{key}_i$, the expected output is the corresponding $\text{value}_i$. This task is a specific type of NIAH test, where the target key-value pair is the "needle" within the $k$ key-value pairs (the "haystack"). 

To generate our synthetic datasets, we create unique and random UUIDs, numbers, or letters to serve as keys and values. For the benchmark Needle-in-a-haystack tests \cite{niah}, the authors prompt GPT-4 to measure the accuracy. However, due to the constraint of accessing GPT-4, we use the ROUGE-1 score \cite{lin2004rouge} to compare the generated text and reference text. For all other key-value retrieval tasks, we measure the exact match accuracy, which assesses if a ground truth value is contained in the model's response.

\subsection{Benchmark tests}
\subsubsection{Key-value retrieval}
For the key-value retrieval benchmark test from \citet{liu2023lost}, we experiment with 500 datasets, each containing 50 key-value pairs of randomly generated UUIDs. While the original benchmark is for 75, 140, and 300 key-value pairs, we limit our datasets to the first 50 pairs from the 75 key-value pairs datasets due to the 4096 input token constraints of GPT-3.5 and LLaMA-2 7B.

\subsubsection{Needle-in-a-haystack retrieval} 
The goal of the Needle-in-a-haystack retrieval task presented by \citet{niah} is to correctly respond to a query based on a statement. The input contexts are essays with a "needle" statement inserted at different depths. To examine the effect of input length on performance, we truncate the essays to various lengths corresponding to 1k, 2k, 3k, and 4k tokens.

\subsection{DENIAHL tasks}

For each DENIAHL task, we utilize 50 randomly generated key-value datasets similar to our examples in Table \ref{tab:example-task}. We calculate the average accuracy scores across these datasets for evaluation. It is important to note that the upper limits for the number of key-value pairs and the length of items are defined by the models' input constraint of 4096 tokens. We aim to maximize the use of the model's capacity without surpassing the limit. Additionally, we examine the retrieval performance by varying the position of the target key-value pair.

The prompt template we use to query the models are inherited from the work of \citet{liu2023lost}. Full examples of our datasets with prompts are shown in Appendix \ref{appendix: examples}.

\subsubsection{Data Size}
When changing the data size by varying the number of items, we test on 10, 30, and 50 key-value pairs, which consist of randomly generated UUIDs as keys and values. 

When changing the data size by modifying the length of items, we vary the length of keys and values to be 8, 16, 32, and 64 characters. UUIDs, which are 32 characters long, are modified for different lengths in our study: for shorter lengths, we extract the first 8 or 16 characters of a UUID; for a length of 64 characters, we concatenate two randomly generated UUIDs.

\subsubsection{Patterns}
For all pattern tasks, we experiment with 100 key-value pairs in the context passed to the LLM given the shorter length of these items. Two numerical patterns and one letter pattern are evaluated.

For numerical patterns, the keys are consecutive integers from 1 to 100. Patterns include one simple pattern and one complex pattern as shown in the Appendix.
\begin{itemize}
    \item The simple pattern follows a global pattern defined by $$\textit{value}_i = \textit{key}_i \cdot \textit{multiplier} + \textit{initial value}.$$ 
    For different datasets, the multiplier and initial value are randomly chosen from values between 1 to 10 and 1 to 20, separately. 
    \item A more complex global pattern is defined by the following. With $\textit{value}_0$ as the $\textit{initial value}$, for $i \geq 1$,
        \begin{align*}
            \textit{value}_i = \left\{
            \begin{aligned}
            &\textit{value}_{i-5} + \textit{increment} \text{, if $i \% 5 = 1$}\\
            &2 \cdot \textit{value}_{i-1} \text{ , otherwise}\\
            \end{aligned}
            \right.
        \end{align*}
    
\end{itemize}

For the letter pattern task, the value is the uppercase of a randomly generated lowercase two-letter key.

\subsubsection{Data Type}
In this task, we experiment with three data types, each characterized by key-value pairs consisting of either entirely random letters, numbers, or a mixture of both. Each dataset comprises 50 key-value pairs, with an item length of 32.

% MENTION THE ABLATION
\section{Results and Discussion}
\subsection{Benchmark Dataset Results}
\subsubsection{Key-value retrieval benchmark}
\begin{figure}[ht!]
\includegraphics[width=70mm]{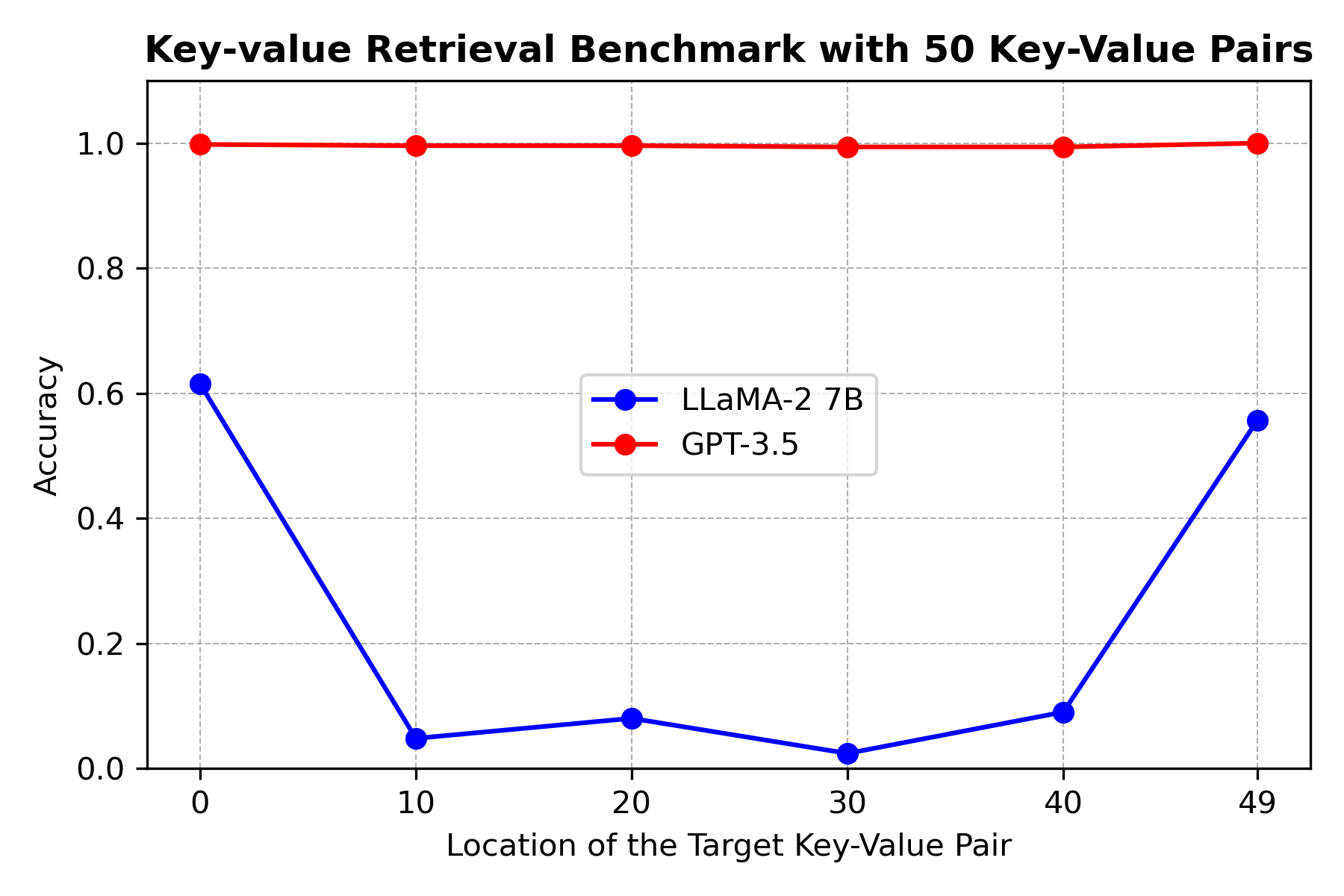}
\caption {LLaMA-2 7B exhibits the "lost-in-the-middle" effect, where accuracy is higher when the target key-value pairs are at the beginning or end of the input context versus in the middle of the context.}
\label {fig: nelson_50kv}
\end{figure}

Figure \ref{fig: nelson_50kv} shows that for LLaMA-2 7B, accuracy is highest when the retrieved values from the key-value pairs are located at the beginning or end of models' input context, demonstrating the lost-in-the-middle effect observed by \citet{liu2023lost} on this benchmark. In general, GPT-3.5 effectively retrieves over its entire context window. 

\subsubsection{Needle-in-a-haystack retrieval benchmark} 
\begin{figure}[ht!]
\centering
\includegraphics[width=60mm]{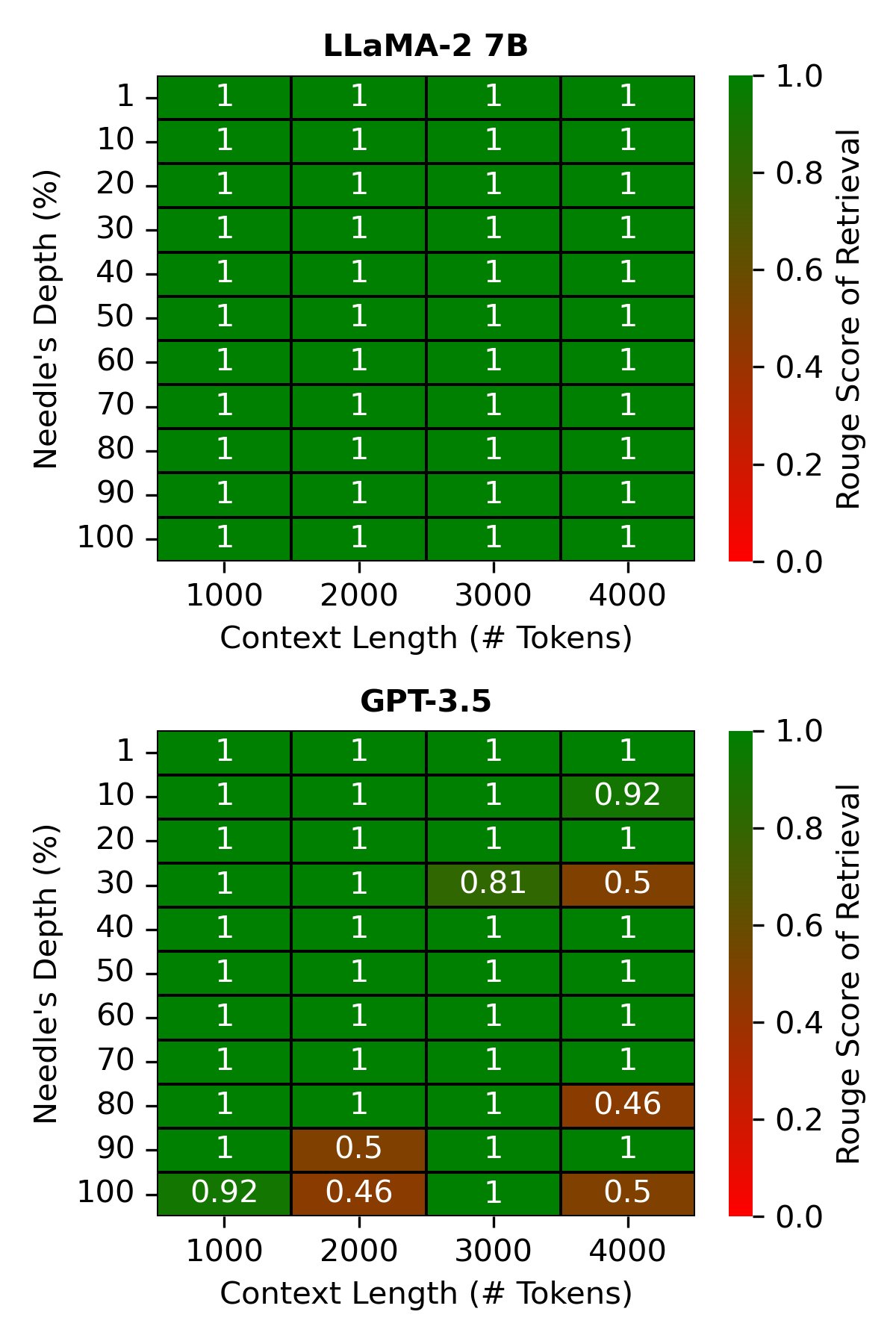}
\caption {ROUGE scores are generally strong for both models on the Needle-in-a-haystack benchmark \cite{niah}.}
\label{fig: niah}
\end{figure}

Figure \ref{fig: niah} demonstrates that both models perform well on the NIAH benchmark \cite{niah}. LLaMA-2 7B appears to outperform GPT-3.5; however, this is due to the ROUGE evaluation metric used. Upon manual examination of responses where GPT-3.5 underperformed, we determine responses are indeed accurate, maintaining 100\% accuracy on the NIAH benchmark. ROUGE-1 scores are lower due to GPT-3.5's tendency to rephrase answers instead of directly replicating them, as in LLaMA-2 7B's output. These effectively perfect results for both models contrasts with LLaMA-2 7B's degraded key-value retrieval benchmark \cite{liu2023lost} which uses random UUIDs. We hypothesize that the better retrieval performance of LLaMA-2 7B on natural language tasks aligns better with the model's training dataset.

\subsection{DENIAHL Results} 
\begin{figure*}[ht!]
\centering
\includegraphics[width=150mm]{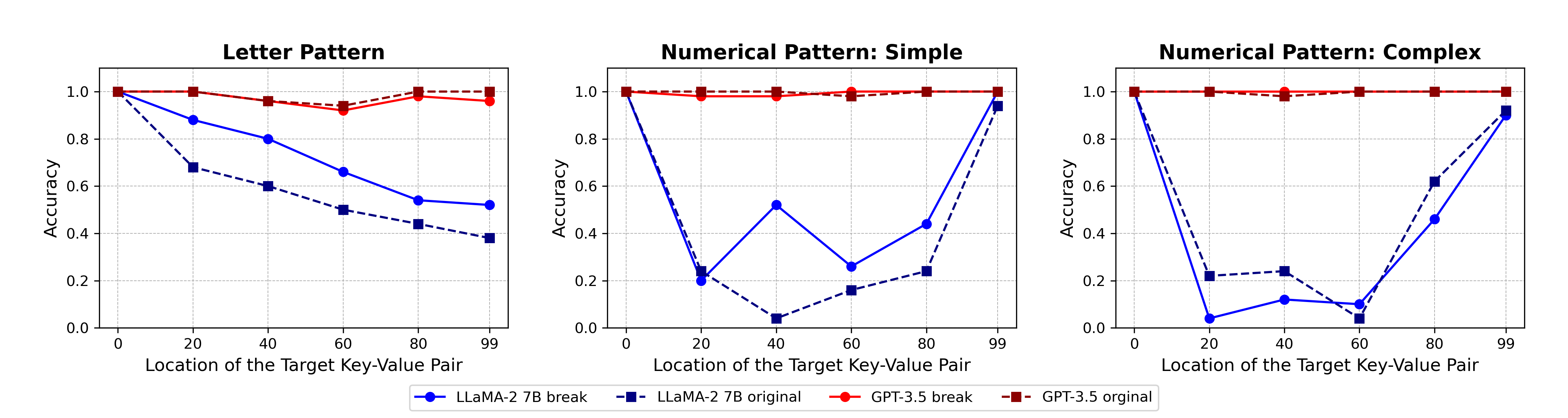}
\caption {Key-value pairs following numerical or letter patterns demonstrate models' preferences for local vs global information}
\label{fig: pattern}
\end{figure*}

In general, as we can see from Figures \ref{fig: item_type}, \ref{fig: data_size}, \ref{fig: pattern},  and \ref{fig: item_type_len}, in most of the tasks, GPT-3.5 achieves nearly a perfect accuracy, while LLaMA-2 7B has more variable performance across the DENIAHL tasks. 

\subsubsection{Data Size} 
\begin{figure*}[ht!]
\centering
\includegraphics[width=150mm]{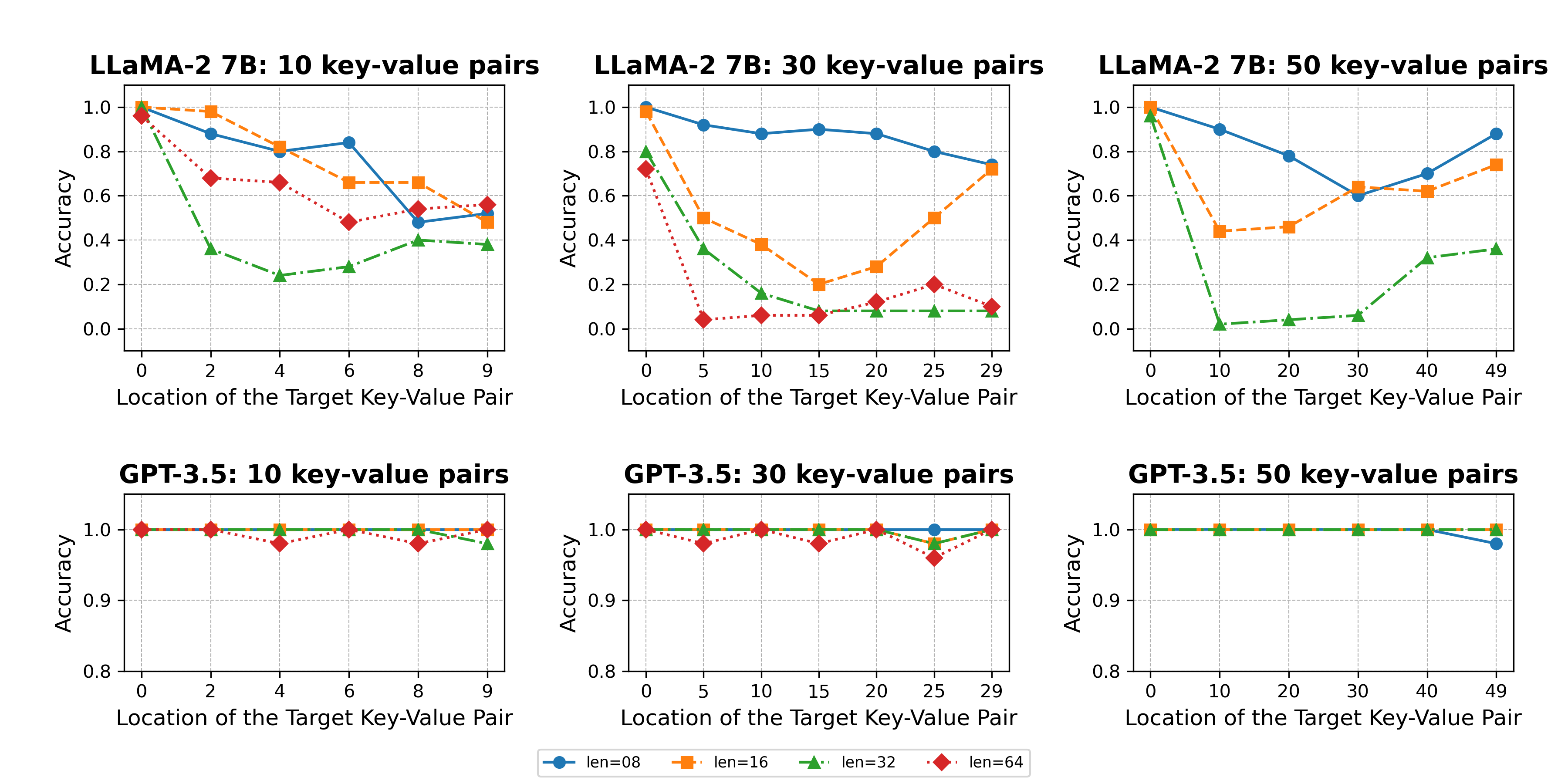}
\caption {Varying data size by changing total number of key-value pairs and number of items influence LLaMA-2 7B performance}
\label{fig: data_size}
\end{figure*}

We find that LLaMA-2 7B generally performs better with 10 key-value pairs compared to 30 or 50 key-value pairs (Figure \ref{fig: data_size}). This suggests that the model more effectively attends to information when the input context length is shorter. 

For the effect of varying the item length, Figure \ref{fig: data_size} shows that recall performance tends to decrease as item length increases. For longer input item lengths of 32 and 64 characters, the performance curves to LLaMA-2 7B have an "L-shape", indicating a "lost-in-the-end" phenomenon. Together, these results suggest recall performance is always strongest at the beginning of an input context regardless of item length, demonstrating models have a bias for primacy over recency.

\subsubsection{Patterns} 
For global pattern attention, we show results for an "original" task where the retrieved value follows the pattern, and "break" where the pattern is broken for the retrieved value. Based on a manual review, when the pattern is broken for LLaMA-2 7B, the model rarely inaccurately inferred the original pattern, instead retrieving the true, altered value. This observation suggests that LLaMA-2 7B does not apply global pattern attention that overrides fine-grained recall capabilities.

Furthermore, as shown in Figure \ref{fig: pattern}, for LLaMA-2 7B, comparing the "break" and "original" modes, we actually observe stronger performance when the pattern is broken for both the letter pattern and the simple numerical pattern. This suggests that the model does not bootstrap global patterns to improve NIAH performance. Instead, the model's local attention bias benefits recall under altered pattern conditions. This finding is however unsupported in the complex numerical pattern task, where the model performs better with the unbroken pattern, although with a smaller difference in performance. For GPT-3.5, the "original" mode shows a slight improvement over the "break" mode in letter patterns, although these differences are far less pronounced than for LLaMA-2 7B.

In addition, for LLaMA-2 7B, the numerical pattern tends to have a "lost-in-the-middle" effect, whereas the letter pattern tends to have a "lost-in-the-end" effect. We further explore how different data types impact model performance in the data type task.

\begin{figure*}[h!]
\centering
\includegraphics[width=150mm]{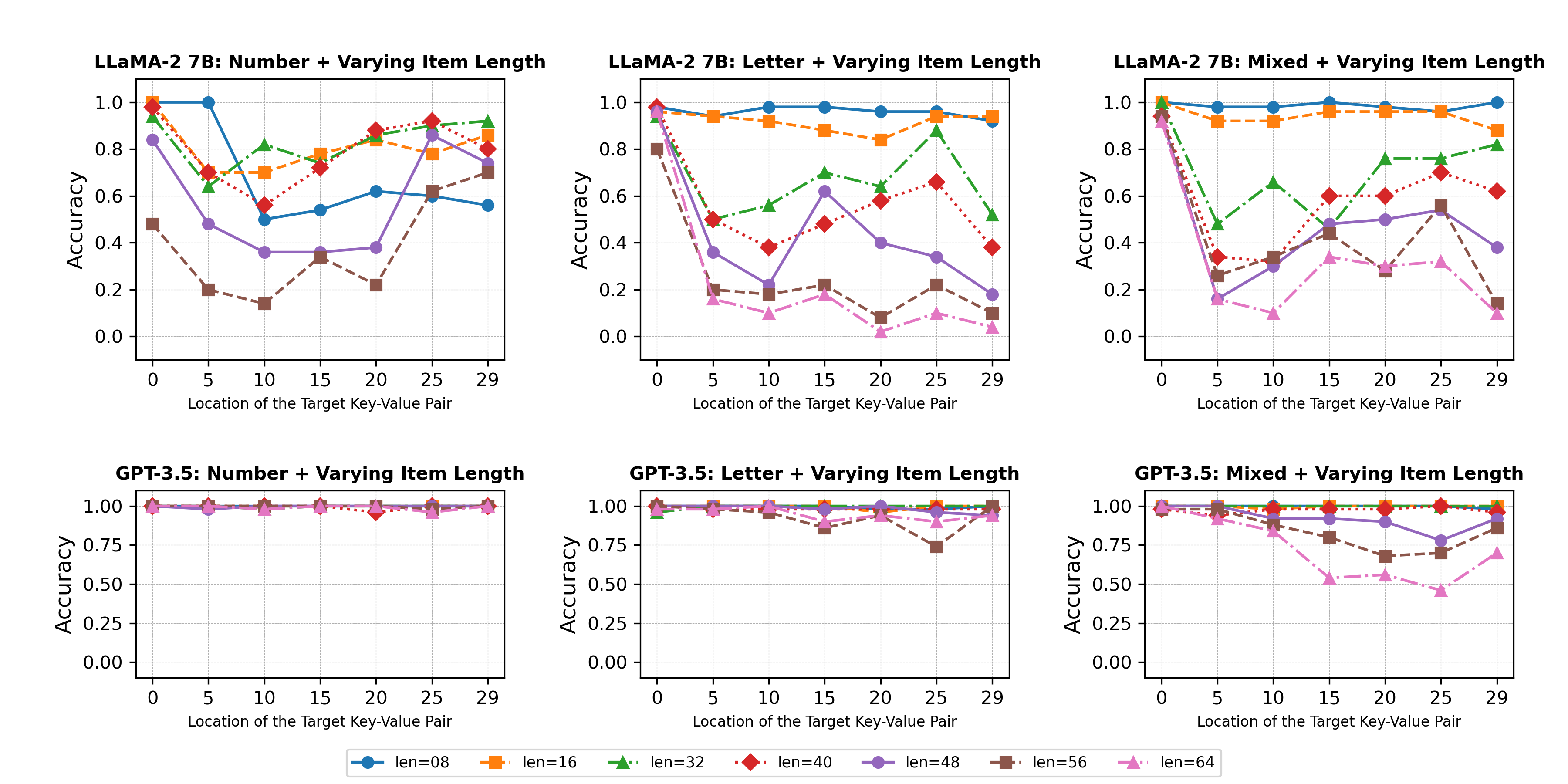}
\caption {Item length and data type in combination change  Needle-in-a-haystack recall capabilities, with longer items decreasing recall performance, and letters and mixed data types proving more difficult to recall.}
\label{fig: item_type_len}
\end{figure*}

\subsubsection{Data Type} 

For LLaMA-2 7B, Figure \ref{fig: item_type} consistently demonstrates a "lost-at-the-end" effect for key-value pairs consisting of random letters and a "lost-in-the-middle" effect for those made up of random numbers. The performance with a mixed data type of numbers and letters follows a similar pattern to that observed with all letters.

To explore whether our findings are applicable across different data sizes, we modify the item length for this task, as illustrated in Figure \ref{fig: item_type_len}. We omit the 64-character length for the numerical data type when running LLaMA-2 7B due to the input length constraint. Generally, increasing input length decreases NIAH recall performance. Our experiment also reveals that for LLaMA-2 7B, longer key-value lengths (48, 56, 64 characters) with the letter data type typically exhibit a "lost-at-the-end" pattern, forming an "L-shape", whereas the number data type generally shows a "U-shape", indicating the "lost-in-the-middle" pattern. The mixed data type displays a more irregular pattern. Thus, when target values are located at the end of the input context, LLaMA-2 7B often performs more accurately with numeric data than with letters. Additionally, letter data shows a wider distribution of recall abilities as the needle location is moved, with primacy biases enabling near perfect recall for early needles even when item length is 64 characters, but dropping to almost 0 at index 20.

Moreover, it is crucial to recognize that we observe noticeable declines in GPT-3.5's performance when key-value pairs reach longer item lengths of 48, 56, and 64 characters, and consist of mixed data types. This suggests that combining data type and item length features can impact GPT-3.5's NIAH abilities, even when either feature in isolation does not.

\section{Conclusions \& Future Work}
We find that Needle-in-a-haystack (NIAH) abilities are influenced by each of the 3 component tasks in DENIAHL: data size, patterns, and type. Whereas some of these features impact recall performance depending on where needles are located in the haystack, other components can alter recall regardless of needle location. In particular, data size is a more globally influential feature, whereas changing data type from numbers to letters can change the emergence of "lost-in-the-middle" into "lost-in-the-end" phenomena, decreasing performance beyond earlier context window positions. This underscores that models' recency biases are subject to change depending on their data input. The whole of these data features (size, type, and patterns) is greater than their sum, with GPT-3.5 only demonstrating reduced NIAH abilities when recalling particularly long mixed data types.

In real world data settings, LLM's are being used to recall information from ever longer contexts. Ensuring that models' input context has data features that make for robust recall is an important initial step before applying these models. Alternatively, leveraging compound AI systems like RAG \cite{compound-ai-blog} for linguistic use cases like \citet{niah} is a promising alternative. Our work also shows that almost all input contexts present strong primacy biases, meaning re-ranking systems like \citet{khattab2020colbert} may similarly prove useful in improving NIAH recall. 

\section{Limitations}

We find the evaluation of long-form text-based NIAH lacking. In our key-value case, we use ROUGE. However, if the generated text rephrases needles (e.g. synonyms), the evaluation won't identify these as correct matches, leading to lower scores despite being technically correct. Therefore, we still need to look for metrics that evaluate extraction other than ROUGE. The original, LLM-as-a-judge metric used by \citet{niah} presents its own limitations. This approach is not only subject to non-deterministic evaluations, but is also unlikely to scale to longer contexts.

Additionally, our NIAH tests may not accurately reflect the extractive capabilities required in differently-structured data in the real world. This data may also have similarly strong influences on NIAH performance, albeit in ways not reflected by our random string manipulation. We may also consider prompt design as an under-analyzed factor of features that influence NIAH performance. Lastly, we only tested 2 models which, although SoTA at the time of writing this paper, still don't reflect the vast array of models that have been tested on other NIAH benchmarks, including those with or without features such as positional encodings that may affect NIAH performance.

\section*{Acknowledgements}
We extend our gratitude to our wonderful course instructor, Sophie Hao, and teaching assistants, Jackson and Cara, for their help throughout this work. We also appreciate the support provided by the NYU High Performance Computing resources.

\bibliography{custom}

\begin{thebibliography}{20}
\expandafter\ifx\csname natexlab\endcsname\relax\def\natexlab#1{#1}\fi

\bibitem[{Anthropic(2024)}]{anthropic2024claude}
Anthropic. 2024.
\newblock The claude 3 model family: Opus, sonnet, haiku.

\bibitem[{Bai et~al.(2023)Bai, Lv, Zhang, Lyu, Tang, Huang, Du, Liu, Zeng, Hou, Dong, Tang, and Li}]{bai2023longbench}
Yushi Bai, Xin Lv, Jiajie Zhang, Hongchang Lyu, Jiankai Tang, Zhidian Huang, Zhengxiao Du, Xiao Liu, Aohan Zeng, Lei Hou, Yuxiao Dong, Jie Tang, and Juanzi Li. 2023.
\newblock \href {http://arxiv.org/abs/2308.14508} {Longbench: A bilingual, multitask benchmark for long context understanding}.

\bibitem[{Bulatov et~al.(2024)Bulatov, Kuratov, Kapushev, and Burtsev}]{bulatov2024scaling}
Aydar Bulatov, Yuri Kuratov, Yermek Kapushev, and Mikhail~S. Burtsev. 2024.
\newblock \href {http://arxiv.org/abs/2304.11062} {Scaling transformer to 1m tokens and beyond with rmt}.

\bibitem[{Ding et~al.(2024)Ding, Zhang, Zhang, Xu, Shang, Xu, Yang, and Yang}]{ding2024longrope}
Yiran Ding, Li~Lyna Zhang, Chengruidong Zhang, Yuanyuan Xu, Ning Shang, Jiahang Xu, Fan Yang, and Mao Yang. 2024.
\newblock \href {http://arxiv.org/abs/2402.13753} {Longrope: Extending llm context window beyond 2 million tokens}.

\bibitem[{Hoffmann et~al.(2022)Hoffmann, Borgeaud, Mensch, Buchatskaya, Cai, Rutherford, de~Las~Casas, Hendricks, Welbl, Clark, Hennigan, Noland, Millican, van~den Driessche, Damoc, Guy, Osindero, Simonyan, Elsen, Rae, Vinyals, and Sifre}]{hoffmann2022training}
Jordan Hoffmann, Sebastian Borgeaud, Arthur Mensch, Elena Buchatskaya, Trevor Cai, Eliza Rutherford, Diego de~Las~Casas, Lisa~Anne Hendricks, Johannes Welbl, Aidan Clark, Tom Hennigan, Eric Noland, Katie Millican, George van~den Driessche, Bogdan Damoc, Aurelia Guy, Simon Osindero, Karen Simonyan, Erich Elsen, Jack~W. Rae, Oriol Vinyals, and Laurent Sifre. 2022.
\newblock \href {http://arxiv.org/abs/2203.15556} {Training compute-optimal large language models}.

\bibitem[{Hsieh et~al.(2024)Hsieh, Sun, Kriman, Acharya, Rekesh, Jia, Zhang, and Ginsburg}]{hsieh2024ruler}
Cheng-Ping Hsieh, Simeng Sun, Samuel Kriman, Shantanu Acharya, Dima Rekesh, Fei Jia, Yang Zhang, and Boris Ginsburg. 2024.
\newblock \href {http://arxiv.org/abs/2404.06654} {Ruler: What's the real context size of your long-context language models?}

\bibitem[{Kamradt(2023)}]{niah}
G.~Kamradt. 2023.
\newblock \href {https://twitter.com/GregKamradt/status/1727018183608193393?lang=en} {Pressure testing claude-2.1 200k via needle-in-a-haystack.}

\bibitem[{Khattab and Zaharia(2020)}]{khattab2020colbert}
Omar Khattab and Matei Zaharia. 2020.
\newblock \href {http://arxiv.org/abs/2004.12832} {Colbert: Efficient and effective passage search via contextualized late interaction over bert}.

\bibitem[{Levy et~al.(2024)Levy, Jacoby, and Goldberg}]{levy2024task}
Mosh Levy, Alon Jacoby, and Yoav Goldberg. 2024.
\newblock \href {http://arxiv.org/abs/2402.14848} {Same task, more tokens: the impact of input length on the reasoning performance of large language models}.

\bibitem[{Li et~al.(2024)Li, Zhang, Do, Yue, and Chen}]{li2024longcontext}
Tianle Li, Ge~Zhang, Quy~Duc Do, Xiang Yue, and Wenhu Chen. 2024.
\newblock \href {http://arxiv.org/abs/2404.02060} {Long-context llms struggle with long in-context learning}.

\bibitem[{Lin(2004)}]{lin2004rouge}
Chin-Yew Lin. 2004.
\newblock Rouge: A package for automatic evaluation of summaries.
\newblock In \emph{Text Summarization Branches Out}, pages 74--81, Barcelona, Spain. Association for Computational Linguistics.

\bibitem[{Liu et~al.(2023)Liu, Lin, Hewitt, Paranjape, Bevilacqua, Petroni, and Liang}]{liu2023lost}
Nelson~F. Liu, Kevin Lin, John Hewitt, Ashwin Paranjape, Michele Bevilacqua, Fabio Petroni, and Percy Liang. 2023.
\newblock \href {http://arxiv.org/abs/2307.03172} {Lost in the middle: How language models use long contexts}.

\bibitem[{Machlab and Battle(2024)}]{machlab2024llm}
Daniel Machlab and Rick Battle. 2024.
\newblock \href {http://arxiv.org/abs/2404.08865} {Llm in-context recall is prompt dependent}.

\bibitem[{Mohtashami and Jaggi(2023)}]{mohtashami2023landmark}
Amirkeivan Mohtashami and Martin Jaggi. 2023.
\newblock \href {http://arxiv.org/abs/2305.16300} {Landmark attention: Random-access infinite context length for transformers}.

\bibitem[{Munkhdalai et~al.(2024)Munkhdalai, Faruqui, and Gopal}]{munkhdalai2024leave}
Tsendsuren Munkhdalai, Manaal Faruqui, and Siddharth Gopal. 2024.
\newblock \href {http://arxiv.org/abs/2404.07143} {Leave no context behind: Efficient infinite context transformers with infini-attention}.

\bibitem[{Pang et~al.(2022)Pang, Parrish, Joshi, Nangia, Phang, Chen, Padmakumar, Ma, Thompson, He, and Bowman}]{pang2022quality}
Richard~Yuanzhe Pang, Alicia Parrish, Nitish Joshi, Nikita Nangia, Jason Phang, Angelica Chen, Vishakh Padmakumar, Johnny Ma, Jana Thompson, He~He, and Samuel~R. Bowman. 2022.
\newblock \href {http://arxiv.org/abs/2112.08608} {Quality: Question answering with long input texts, yes!}

\bibitem[{Wu et~al.(2024)Wu, Wang, Xiao, Peng, and Fu}]{wu2024retrieval}
Wenhao Wu, Yizhong Wang, Guangxuan Xiao, Hao Peng, and Yao Fu. 2024.
\newblock \href {http://arxiv.org/abs/2404.15574} {Retrieval head mechanistically explains long-context factuality}.

\bibitem[{Xiong et~al.(2023)Xiong, Liu, Molybog, Zhang, Bhargava, Hou, Martin, Rungta, Sankararaman, Oguz, Khabsa, Fang, Mehdad, Narang, Malik, Fan, Bhosale, Edunov, Lewis, Wang, and Ma}]{xiong2023effective}
Wenhan Xiong, Jingyu Liu, Igor Molybog, Hejia Zhang, Prajjwal Bhargava, Rui Hou, Louis Martin, Rashi Rungta, Karthik~Abinav Sankararaman, Barlas Oguz, Madian Khabsa, Han Fang, Yashar Mehdad, Sharan Narang, Kshitiz Malik, Angela Fan, Shruti Bhosale, Sergey Edunov, Mike Lewis, Sinong Wang, and Hao Ma. 2023.
\newblock \href {http://arxiv.org/abs/2309.16039} {Effective long-context scaling of foundation models}.

\bibitem[{Xu et~al.(2024)Xu, Ping, Wu, McAfee, Zhu, Liu, Subramanian, Bakhturina, Shoeybi, and Catanzaro}]{xu2024retrieval}
Peng Xu, Wei Ping, Xianchao Wu, Lawrence McAfee, Chen Zhu, Zihan Liu, Sandeep Subramanian, Evelina Bakhturina, Mohammad Shoeybi, and Bryan Catanzaro. 2024.
\newblock \href {http://arxiv.org/abs/2310.03025} {Retrieval meets long context large language models}.

\bibitem[{Zaharia et~al.(2024)Zaharia, Khattab, Chen, Davis, Miller, Potts, Zou, Carbin, Frankle, Rao, and Ghodsi}]{compound-ai-blog}
Matei Zaharia, Omar Khattab, Lingjiao Chen, Jared~Quincy Davis, Heather Miller, Chris Potts, James Zou, Michael Carbin, Jonathan Frankle, Naveen Rao, and Ali Ghodsi. 2024.
\newblock The shift from models to compound ai systems.
\newblock \url{https://bair.berkeley.edu/blog/2024/02/18/compound-ai-systems/}.

\end{thebibliography}
\onecolumn
\newpage
\appendix
% Can switch back to two-column layout
% Uncomment this if needed \twocolumn

\section*{Appendix}
\section{Prompt Examples}
\label{appendix: examples}

\begin{itemize}
\item \textbf{Data Size: Number of Items}

Number of key-value pairs = 5
\begin{tcolorbox}
Extract the value corresponding to the specified key in the JSON object below.

JSON data:
\begin{verbatim}
{"1812b65a-6db8-48a3-b126-9f18dfe2": "da168bac-843b-4df2-9f80-b38ddd56",
 "852d943e-ac08-418a-a66b-cbccc6aa": "6dea2f5b-dbdd-4386-b8c7-9694902f",
 "e048025a-e85a-4219-9643-5c30900e": "a5874a61-3c0f-4d79-b131-2c2176f8",
 "bf92bd2c-454c-48f3-b1f0-7b56233b": "5e768f0e-5c29-4ac0-ad44-c876c5f3",
 "85f98907-18fa-45f1-8ee6-2db794bc": "f3935e8f-3402-4da3-9976-f6c06b8b"
}
\end{verbatim}
Key: "e048025a-e85a-4219-9643-5c30900e"

Corresponding value:
\end{tcolorbox}

\item \textbf{Data Size: Item Length}

Item length = 8
\begin{tcolorbox}
Extract the value corresponding to the specified key in the JSON object below.

JSON data:
\begin{verbatim}
{"66b1b1c9": "d53e0cf5",
 "6669d056": "cae5c7ee",
 ...
 "9d1e4c53": "25f49ad6"}
\end{verbatim}
Key: "66b1b1c9"

Corresponding value:
\end{tcolorbox}
Item length = 16
\begin{tcolorbox}
Extract the value corresponding to the specified key in the JSON object below.

JSON data:
\begin{verbatim}
{"3ed69dd2-7c6d-4d": "94743764-dd9d-48",
 "f6c8a039-52ee-4d": "51bca270-559a-42",
 ...
 "304d6216-9aca-4d": "901368b1-8229-48"}
\end{verbatim}
Key: "f6c8a039-52ee-4d"

Corresponding value:
\end{tcolorbox}

\item \textbf{Pattern}

Numerical Pattern

\begin{table}[ht]
\centering
\begin{tabular}{cccc}
\toprule
\multicolumn{2}{c}{Simple Pattern} & \multicolumn{2}{c}{Complex Pattern} \\
\cmidrule(lr){1-2} \cmidrule(lr){3-4}
Key & Value & Key & Value \\
\midrule
1 & 6  & 1 & 3 \\
2 & 12 & 2 & 6 \\
3 & 18 & 3 & 12 \\
4 & 24 & 4 & 24 \\
5 & 30 & 5 & 48 \\
6 & 36 & 6 & 5 \\
7 & 42 & 7 & 10 \\
\ldots & \ldots & \ldots & \ldots \\
\midrule
\multicolumn{2}{c}{Initial value = 6} & \multicolumn{2}{c}{Initial value = 3} \\
\multicolumn{2}{c}{ Multiplier = 2} & \multicolumn{2}{c}{Increment = 2} \\
\bottomrule
\end{tabular}
\caption{Examples of two numerical patterns.}
\label{tab:pattern-example}
\end{table}

\begin{tcolorbox}

Extract the value corresponding to the specified key in the JSON object below.

JSON data:
\begin{verbatim}
{"1": "6",
 "2": "12",
 ...
 "10": "60"}
\end{verbatim}
Key: "10"

Corresponding value:
\end{tcolorbox}
Letter Pattern
\begin{tcolorbox}
Extract the value corresponding to the specified key in the JSON object below.

JSON data:
\begin{verbatim}
{"xf": "XF",
 "qh": "QH",
 ...
 "pr": "PR"}
\end{verbatim}
Key: "pr"

Corresponding value:
\end{tcolorbox}

% Mixed Pattern
% \begin{tcolorbox}

% Extract the value corresponding to the specified key in the JSON object below.

% JSON data:
% \begin{verbatim}
% {0: "AA",1: "BB",2: "CC",
% 3: "DD",4: "XX",5: "FF"}
% \end{verbatim}
% Key: 4

% Corresponding value:
% \end{tcolorbox}

\item \textbf{Data Type}

All Letters
\begin{tcolorbox}
Extract the value corresponding to the specified key in the JSON object below.

JSON data:
\begin{verbatim}
{"cayZasqdCrhmfjveShWsKKSMgKqGzZeX": "sNTbeZuyULXsibrYxfPjnjcztHXyJDAO",
"vamEFQdrsUUiwzEaqmCrlzsrhvbotsFv": "TpxEXKJTDsHgGXOSWXLuNDqRTYoyweoo",
...
"yBeOhaxFlchqcoWZtZSUFyWwankqDiFU": "zSfhwImONwgAxEaaeKDIDRbZeOYsvOSD"}
\end{verbatim}
Key: "vamEFQdrsUUiwzEaqmCrlzsrhvbotsFv"

Corresponding value:
\end{tcolorbox}
All Numbers
\begin{tcolorbox}
Extract the value corresponding to the specified key in the JSON object below.

JSON data:
\begin{verbatim}
{"53686654614704676755476453368009": "79478310318543681189996595410008",
"93617318386685631296611796626551": "54636111463827476966944487807192",
...
"49409497339768004959989982675292": "41860753494712594864088858418453"}
\end{verbatim}
Key: "53686654614704676755476453368009"

Corresponding value:
\end{tcolorbox}
Mixed (Letters and Numbers)
\begin{tcolorbox}
Extract the value corresponding to the specified key in the JSON object below.

JSON data:
\begin{verbatim}
{"f97snUuLlUEdxjPONaas9DTcoIKxLYcC": "RJWo8ulz0LMA1MDKlhMUI5mA8QARv2il",
"y9oCFKKx9Kbaem4IsaFTIQaSrBnD6KFi": "g6yZ5KhgByrFHyPwpTcbnm9rMga8NK8V",
"9iwoh7ARUCjodT26ZYUzLhyg8LvqXtk2": "W0m0iG25QgyAArO1uiKmuZLMbfgzbLu5"}
\end{verbatim}
Key: "9iwoh7ARUCjodT26ZYUzLhyg8LvqXtk2"

Corresponding value:
\end{tcolorbox}
\end{itemize}

\end{document}